\documentclass[11pt,letterpaper]{article}
\usepackage[hyperref]{acl2020}
\usepackage{times}
\usepackage{latexsym}

\usepackage{microtype}
\usepackage{enumitem}
\usepackage{algorithm}
\usepackage[noend]{algpseudocode}

\usepackage{adjustbox}
\usepackage{multirow}
\usepackage{bbm}
\usepackage{textcomp}
\usepackage{amsmath}
\usepackage{resizegather}
\usepackage{verbatim,multirow,mathtools,enumitem,textcomp,capt-of,url,color,lipsum,epstopdf,graphicx,relsize,balance,times,layouts,amssymb}
\usepackage{booktabs}
\usepackage{xcolor}

\newcommand{\sysname}{\textsc{ZSCeres}}
\newcommand{\newwebsite}{Unseen-Website}
\newcommand{\newvertical}{Unseen-Vertical}

\aclfinalcopy 
\def\aclpaperid{2427} 

\title{ZeroShotCeres: Zero-Shot Relation Extraction \\ from Semi-Structured Webpages}

\author{
 Colin Lockard \\
 University of Washington \\
  \texttt{ lockardc@cs.washington.edu} \\
\And
 Prashant Shiralkar\\
 Amazon \\
  \texttt{shiralp@amazon.com} \\
\AND
Xin Luna Dong \\
Amazon\\
 \texttt {lunadong@amazon.com}\\
\And
 Hannaneh Hajishirzi \\
 University of Washington, 
 Allen Institute for AI \\
  \texttt{hannaneh@cs.washington.edu} \\ 
}

\date{}

\begin{document}
\maketitle
\begin{abstract}

In many documents, such as semi-structured webpages, textual semantics are augmented with additional information conveyed using visual elements including layout, font size, and color. Prior work on information extraction from semi-structured websites has required learning an extraction model specific to a given template via either manually labeled or distantly supervised data from that template. In this work, we propose a solution for ``zero-shot'' open-domain relation extraction from webpages with a previously unseen template, including from websites with little overlap with existing sources of knowledge for distant supervision and websites in entirely new subject verticals. Our model uses a graph neural network-based approach to build a rich representation of text fields on a webpage and the relationships between them, enabling generalization to new templates. Experiments show this approach provides a 31\% F1 gain over a baseline for zero-shot extraction in a new subject vertical.

\end{abstract}

\section{Introduction}

Semi-structured websites offer rich sources of high-quality data across many areas of knowledge \cite{Dong2014KnowledgeVA}. These websites present information via text that is accompanied by rich visual and layout features that can be generalized beyond a single website.   However, most prior work on information extraction (IE) from  websites has largely ignored most of these features, instead relying only on HTML features specific to an individual website \cite{Ferrara2014WebDE}. This requires training data for every website targeted for extraction, an approach that cannot scale up if training data must be manually created.

\begin{figure}[t]
  \centering
    \includegraphics[width=1.0\linewidth,scale=1]{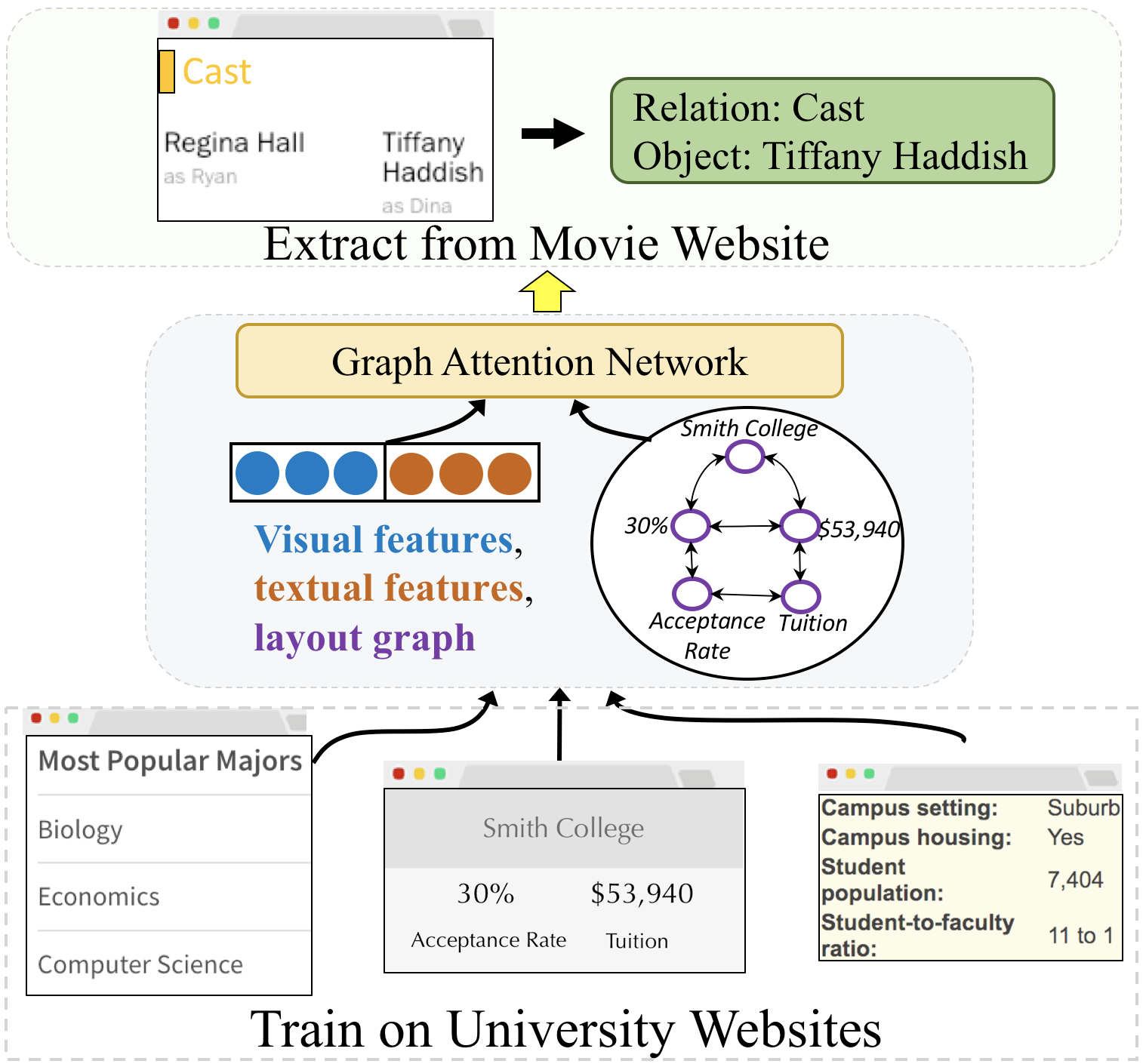}
  \caption{Our zero-shot open-domain information extraction process learns generalizable graph-based representations of how relations are visually presented on semi-structured websites, allowing for training on one vertical (such University sites) and extraction from another (such as Movie sites).}
  \label{fig:teaser}
  \vspace{-0.5em}
\end{figure}

To circumvent manual data annotation, previous work used a distant supervision process requiring a knowledge base aligned to the website targeted for extraction \cite{Gentile2015EarlyST, Lockard2018CERESDS}, including for OpenIE extraction \cite{Banko2007OpenIE, Bronzi2013ExtractionAI, Lockard2019OpenCeresWO}.  
These methods, however, can only learn a website-specific model based on seed knowledge for the site, but cannot be generalized to the majority of websites with knowledge from new verticals, by long-tail specialists, and in different languages.

In this paper, we introduce the task of {\em zero-shot relation extraction} from semi-structured websites, in which a learned model is applied to extract from a website that was not represented in its training data (Figure~\ref{fig:teaser}). Moreover, we introduce \textsc{ZeroShotCeres}, a graph neural network model that  encodes  semantic textual and visual patterns common across different training websites and  can generalize to extract information from documents with never-before-seen templates and topics.

Unlike unstructured text, which can be modeled as a sequence, or images, which can be modeled as a two-dimensional grid of pixels, it is not obvious how to operate over the many shapes and sizes of text fields on a semi-structured webpage. We illustrate our intuition using the webpage snippets in Figure \ref{fig:teaser}: Despite their differences, each site uses alignment of relation and object strings, either vertically or horizontally, to help indicate relationships; in addition, relation strings are often more prominent than their objects, either in size or boldness. 
Such features are semantically meaningful to readers and often consistent from site to site; thus, encoding them into the representation of webpages will allow us to generalize to unseen sites.

Our model, \textsc{ZeroShotCeres}, encodes these diverse feature types in a graph representation in which each text field becomes a node in a graph, connected by edges indicating layout relationships on the page. This abstracts away the details of the page while maintaining the core visual structure presented to the reader. A graph neural network is then applied to produce a new representation of each text field, informed by the surrounding page context. This representation is then used to extract entities and relationships from the document. This allows us to extract not only in the closed-domain setting, but also allows us to conduct OpenIE on websites about entirely new subject verticals not seen during training.

Our  contributions are threefold: 
    (a) We introduce a graph neural network model for webpage representation that integrates multi-modal information including visual, layout, and textual features, enabling generalization for IE from never-before-seen websites.
 (b) We propose the first approach to enable Open Information Extraction from semi-structured websites without prior knowledge or training data in the subject vertical.
  (c) Our method works in both OpenIE and ClosedIE settings. We conduct evaluations showing the effectiveness of the technique and exploring the challenges of zero-shot semi-structured IE, achieving a 31\% improvement in F1 compared to an OpenIE baseline. The graph model gives a 26\% F1 boost when extracting according to a defined schema (ClosedIE).

\section{Related Work}
\label{sec:related_work}

\noindent \textbf{DOM-based ClosedIE:}
The conventional approach to extraction from semi-structured websites is wrapper induction \cite{Kushmerick1997WrapperIF}, in which training data for documents from a given template is used to learn a rule-based extractor based on DOM (i.e., HTML) features to apply to other documents of the same template, extracting relations according to a pre-defined ontology (``ClosedIE''). Since this approach requires training data for each template targeted for extraction, recent work has focused on reducing the manual work needed per site. Fonduer \cite{Wu2018FonduerKB} provides an interface for easily creating training data, Vertex \cite{Gulhane2011WebscaleIE} uses semi-supervision to minimize the number of labels needed, LODIE \cite{Gentile2015EarlyST} and Ceres \cite{Lockard2018CERESDS} automatically generate training data based on distant supervision, and DIADEM \cite{Furche2014DIADEMTO} identifies matching rules for specific entity types.

\begin{figure*}[t]
  \centering
    \includegraphics[width=.9\linewidth]{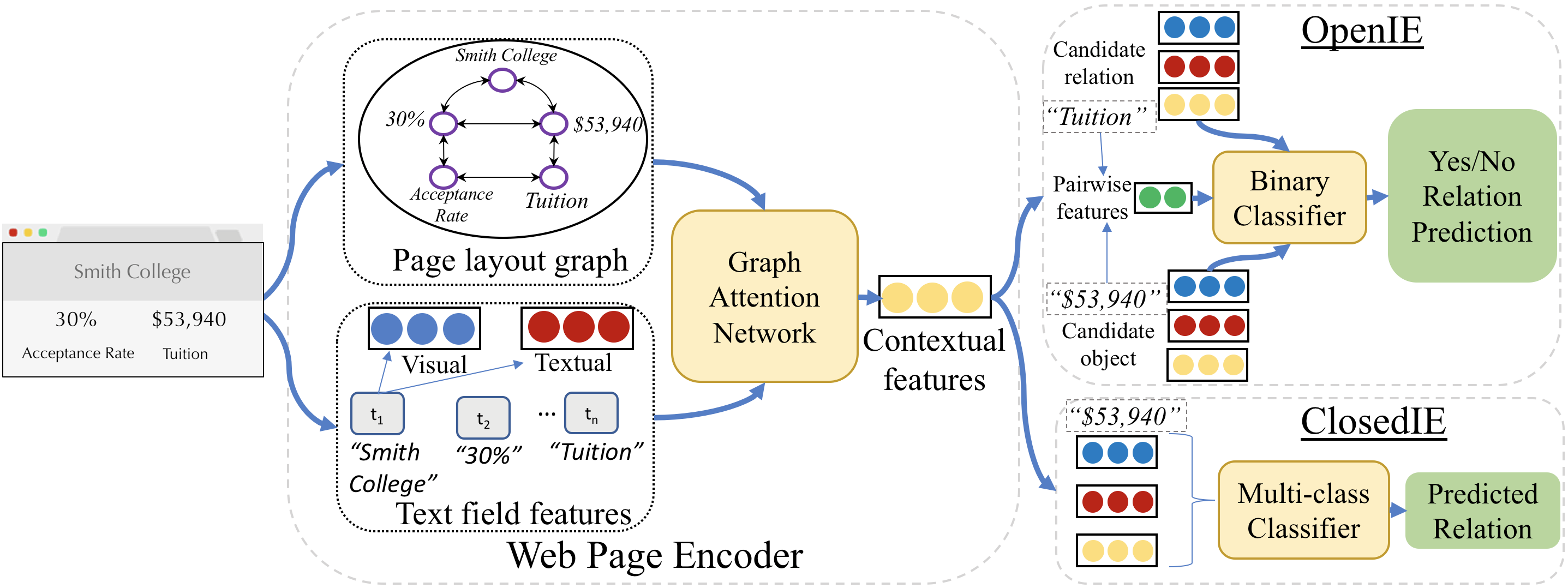}
  \caption{A depiction of the web page representation module (left) and relation classifiers (right).} 
  \label{fig:model_diagram}
  \vspace{-0.5em}
\end{figure*}

\smallskip
\noindent \textbf{DOM-based OpenIE:} WEIR \cite{Bronzi2013ExtractionAI} and OpenCeres \cite{Lockard2019OpenCeresWO} offer OpenIE approaches to DOM extraction. The latter method uses visual features in a semi-supervised learning setting to identify candidate pairs that are visually similar to known \textit{(relation, object)} pairs; however, the ultimate extraction model learned is still site-specific and based on DOM features rather than the more generalizable visual or textual features. \citet{Pasupat2014ZeroshotEE} present a zero-shot method for extraction from semi-structured webpages, but limit their work to extraction of entities rather than relationships and do not consider visual elements of the page.

\smallskip
\noindent \textbf{Multi-modal extraction:}
The incorporation of visual information into IE was proposed by \citet{Aumann2006VisualIE}, who attempted to learn a fitness function to calculate the visual similarity of a document to one in its training set to extract elements like headlines and authors. Other recent approaches that attempt to address the layout structure of documents are CharGrid \cite{Katti2018ChargridTU}, which represents a document as a two-dimensional grid of characters, RiSER, an extraction technique targeted at templated emails \cite{Kocayusufoglu2019RiSERLB}, and that by \citet{Liu2018HierarchicalRF}, which presents an RNN method for learning DOM-tree rules. However, none of these address the OpenIE setting, which requires understanding the relationship between different text fields on the page. 

The approaches most similar to ours are GraphIE \cite{Qian2018GraphIEAG} and the approach by \citet{Liu2019GraphCF}. Both approaches involve constructing a graph of text fields with edges representing horizontal and vertical adjacency, followed by an application of a GCN. However, neither approach makes use of visual features beyond text field adjacency nor DOM features, and both only consider extraction from a single text field rather than OpenIE. In addition, they show only very limited results on the ability of their model to generalize beyond the templates present in the training set.

\section{Problem  and Approach Overview}
\label{sec:problem_definition}

\subsection{Zero-shot relation extraction from semi-structured websites}
We address the problem of extracting entities and
the relationships between them as expressed by never-before-seen
semi-structured websites. A {\it semi-structured website} typically belongs to a subject {\it vertical} $V$, where $V$ is a general field of knowledge such as movies, finance, or sports.
 A semi-structured website consists of a set of \textit{detail pages} sharing a similar template, each of which contains a set of facts about a page topic entity $e_{topic}$. The HTML document $w$ defines a set of text fields $T$, which the web browser renders as a \textit{webpage} according to the instructions defined in the HTML and any referenced auxiliary files such as CSS or Javascript. The text fields have both textual and visual features, described in Section \ref{sec:text_field_encoding}.

\subsubsection{Relation Extraction}

Our goal is to extract \textit{(subject, relation, object)} knowledge triples, where the subject is $e_{topic}$, the object is a text field $t \in T$ containing the name of an entity (or atomic attribute value), and the relation indicates the relationship between the two entities. 

For this work, we assume the page topic entity has already been identified, (such as by the method proposed by \citet{Lockard2018CERESDS} or by using the HTML \texttt{title} tag) and thus limit ourselves to identifying the objects and corresponding relations. We consider the following two settings:

\vspace{.1cm}
\noindent \textbf{Relation Extraction (ClosedIE):} Let $R$ define a closed set of relation types, including a special type indicating ``No Relation''. Relation Extraction is the assignment of each text field $t$ to one $r_i \in R$, which indicates the relationship between the entity $e_{object}$ mentioned in $t$ and $e_{topic}$.  

\vspace{.1cm}
\noindent \textbf{Open Relation Extraction (OpenIE):} Given a pair of text fields $(i, j)$, Open Relation Extraction is a binary prediction of whether $i$ is a relation string indicating a relationship between the entity $e_{object}$ mentioned in $j$ and $e_{topic}$. 

\subsubsection{Zero-shot Extraction}

Unlike prior work that requires the learning of a model specific to the semi-structured website targeted for extraction, we look at zero-shot extraction.
 Given a semi-structured website $W$ targeted for extraction, zero-shot extraction is the learning of a model without any use of pages from $W$ during training.
We consider two zero-shot settings:

\vspace{.1cm}
\noindent \textbf{{\newwebsite} Zero-shot Extraction} is the learning of a model without any use of pages from $W$, but with pages from some other website(s) from vertical $V$ during training.

\vspace{.1cm}
\noindent \textbf{{\newvertical} Zero-shot Extraction} is the learning of a model without any use of pages from $W$ or of pages from any website with vertical $V$ during training.

\subsection{Approach Overview} 
Figure~\ref{fig:model_diagram} depicts our approach for zero-shot relation extraction (detailed in Section~\ref{sec:relation_extraction_model}) leveraging  a web page representation that will capture the similarities in visual and textual semantics across websites (Section~\ref{sec:web_page_encoder}). Our web page representation  module first converts each page into a layout graph (Section~\ref{sec:page_graph_construction}) that abstracts away the details of the page structure while maintaining the adjacency relationships between text fields. We represent each text field with an initial feature vector of visual and textual attributes. 
This input is passed into a graph neural network that allows for information to flow between nodes, producing a new text field representation that captures contextual information (Section~\ref{sec:gnn}). 

To obtain a web page encoding, we leverage a pre-training step with auxilliary loss function $\mathcal{L}_{pre}$ that encourages the model to produce an intermediary representation useful for IE. This is performed via a three-way classification that determines if a text field contains a relation name, the object of some relation, or irrelevant text (Section~\ref{sec:pretraining}). After pre-training, the weights of this GNN are frozen and it can be applied to new pages, with its output used as input into a relation extraction module, optimized with task-specific loss function $\mathcal{L}_{task}$, where the task is either OpenIE or ClosedIE, described in Section \ref{sec:relation_extraction_model}. 
 The resulting approach minimizes our overall loss $\mathcal{L}_{\sysname}$, with:
 \begin{equation}
 \mathcal{L}_{\sysname} = \mathcal{L}_{pre} + \mathcal{L}_{task} 
 \end{equation}

\section{Web Page Encoder}
\label{sec:web_page_encoder}

The key idea behind our solution is to train webpage representations to capture the fundamental similarities in visual and textual semantics across websites to express relations, objects, and their relationships. The fundamental characteristics we capture, generalizable across templates and verticals, thus allow us to carry over our knowledge across websites and enable zero-shot extraction.

There are two key parts in our solution. First, we build a graph to capture the layout relationships in a more abstract form that allows us to more easily learn the common features across different sites such as the fact that relation strings are often to the left or above their objects (Section \ref{sec:page_graph_construction}). 
Second, we apply a Graph Neural Network (GNN) to learn 
representations for each node capturing contextual information about its neighborhood on the webpage (Section \ref{sec:gnn}), allowing information to flow through the nodes, providing context (e.g., flowing through ``Cast'' to a far-away node ``Uma Thurman'' via the closer node ``Ethan Hawke'' in Figure \ref{fig:example_page_with_graph}). 
This representation will be useful for relation extraction as described in Section \ref{sec:relation_extraction_model}.

\subsection{Page graph construction}
\label{sec:page_graph_construction}
We encode the layout relationships between text fields in the form of a graph, $G$, consisting of a set of nodes $N$, each corresponding to a text field, and a set of edges $E$ corresponding to relationships between the text fields. The edges capture three forms of adjacency, as shown in the example in Figure \ref{fig:example_page_with_graph}: 

\begin{figure}[t]
  \centering
    \includegraphics[width=0.9\linewidth,scale=1]{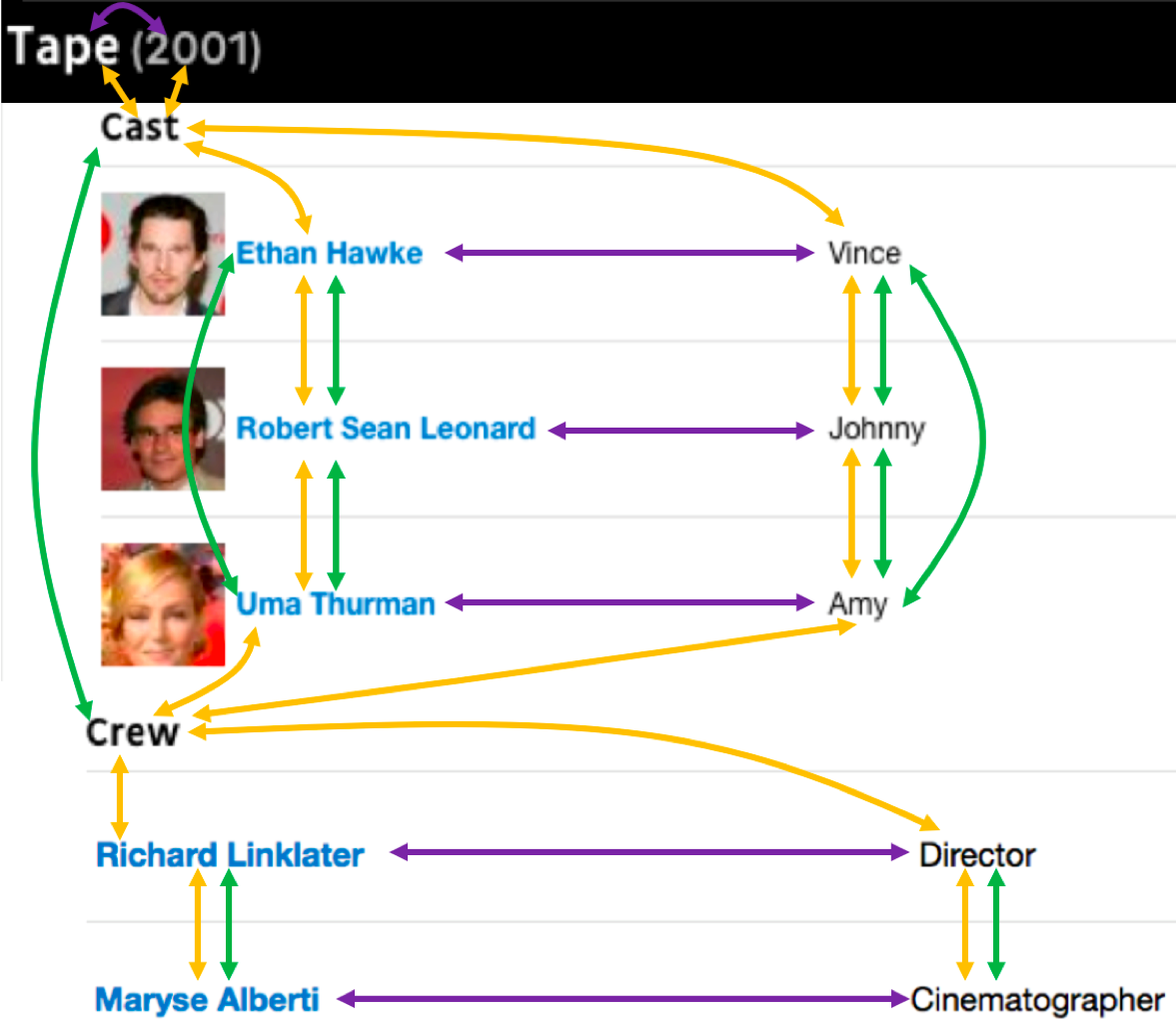}
  \caption{A cropped portion of the detail page from allmovie.com for the film \emph{Tape}. Arrows overlaid showing the constructed page graph consisting of edges for each horizontal (purple), vertical (yellow) and DOM (green) relationship between text fields. }
  \label{fig:example_page_with_graph}
  \vspace{-0.5em}
\end{figure}

\vspace{.1cm}
\noindent \textbf{Horizontal:} Edges are added when two text fields are horizontal neighbors on the page; that is, they have a shared vertical location and there are no other text fields between them.

\vspace{.1cm}
\noindent \textbf{Vertical:} Edges are added when two text fields are vertical neighbors on the page; that is, they have an overlapping horizontal location and there are no other text fields between them.

\vspace{.1cm}
\noindent \textbf{DOM:} Edges are added when two text fields are siblings or cousins in the DOM tree; that is, the absolute XPaths identifying their locations differ only at a single index value.

\subsection{Graph Neural Network (GNN)}
\label{sec:gnn}
To build a representation of each text field that incorporates the surrounding page context, we use Graph Attention Networks (GAT) \cite{velickovic2018graph}. The feature vector for each text field (described below) and the page graph form the input to a GAT, which then produces a new representation for each text field based on the surrounding context in the graph. Specifically, for each text field $i$, GAT layer $l$ computes a representation $ h_i^l$ as follows:
\begin{equation}
 h^{l}_i = \sigma \Bigg( \sum_{j \in N_i} {\alpha_{ij} W^{l}_{G}h^{l-1}_j} \Bigg), 
\end{equation}

\noindent where $N_i$ is the set of neighbors of node $i$ in the graph, and $h_j^{l-1}$ is the representation of node $j$ from the preceding layer; $h_j^0$ indicates the input features for the node. (For each node, we add a self loop to the graph; that is, including $i$ in $N_i$.) $W^{l}_{G}$ is a learned weight matrix applied to the node features for layer $l-1$ and $\sigma$ is a non-linear function, in our case a ReLU. The attention weight $\alpha_{ij}$ determines how influenced a node's representation is by each of its neighbors, calculated as follows:
\begin{equation}
\alpha_{ij} = \frac {\exp{\Big(\sigma\big(a^\top [W^{l}_{G}h^{l-1}_i;W^{l}_{G}h^{l-1}_j}]\big)\Big)} {\sum_{k \in N_i} {\exp{\Big(\sigma\big(a^\top [W^{l}_{G}h^{l-1}_i;W^{l}_{g}h^{l-1}_k}]\big)\Big)}} ,
\end{equation}
\noindent where $a$ is a weight vector applied against the concatenation (represented by ``$;$'') of the two node's features as transformed by $W^{l}_{G}$ and $\sigma$ is a ReLU.  This produces a new contextualized set of features for each node that are informed by the surrounding page context. We describe the original input features for each text field in the next section.

\subsubsection{Initial text field features} 

\label{sec:text_field_encoding}
For each text field  on the page, we produce an initial feature vector containing both visual feature vector $V$  and textual feature vector $T$. We define the input feature vector $h^0_i$ for text field $i$ as: 

\begin{equation}
h^0_i = [T(i);V(i)]    
\end{equation}

\noindent where ``$;$'' represents concatenation. 

\vspace{.1cm}
\noindent \textbf{Visual Features:} A numeric feature vector is constructed representing the bounding box coordinates of the text field, the height and width of the bounding box, and the font size, along with one-hot features representing the typeface, font weight, font style, color, and text alignment.

\vspace{.1cm}
\noindent \textbf{Textual Features:} In ClosedIE, to capture its semantics, the textual content of the text field is processed with a pre-trained BERT \cite{Devlin2018BERTPO} model. To produce a representation of the entire text field, we simply average the BERT-Base output for each token in the text field. For OpenIE, since the goal is to generalize to entirely new subject verticals that may contain text not seen during training, only a single textual feature is used\footnote{This feature is also used during ClosedIE}: the percent of pages on the site on which the string in the text field appears. This frequency measure helps differentiate relation strings, which are likely to be common, from object strings, which are more likely to be rare.

\subsection{Pre-Training Web Page Encoder}
\label{sec:pretraining}
To encourage the GNN weights to capture the features necessary to represent relationships on the page, we use a pre-training step to learn the GNN representation before incorporating it into the extraction model. 
The pre-training task is a simplified form of the OpenIE task. To speed up training by avoiding the pairwise decisions necessary for OpenIE, we instead perform a multi-class classification of each text field into a class $c$ in the set \{Relation, Object, Other\}: 
\begin{equation}
 p\left(c | h^l_i;\theta\right)  = softmax \left( W_{pre} h_i^l \right)    
\end{equation}
\noindent where $h_i^l$ is the output of the GNN for the text field, $W_{pre}$ is a weight matrix, and $\theta$ comprises $W_G$ and $W_{pre}$. Given a training set with $T$ text fields, each with a ground truth class $y^{pre}_i$, we minimize the cross-entropy loss $\mathcal{L}_{pre}$:
\begin{equation}
 \mathcal{L}_{pre} = - \sum_{i=1}^{T} \log  p \left( y^{pre}_i | h^l_i, \theta \right)
\end{equation}
To discourage overfitting to spurious details in the small number of websites in our training set, we freeze the GNN weights after pre-training and do not update them during the full OpenIE training. After pre-training we discard the linear layer $W_{pre}$ since it is not needed for subsequent steps; instead, we directly use the GNN output $h^l$.

\section{Relation Extraction Model}
\label{sec:relation_extraction_model}
Once we have the new representation $h_t^l$ of each text field $t$ produced by the above GNN process, we can perform our final classification. 

\subsection{OpenIE}
For OpenIE, the classification decision must be made over a pair of text fields, $i$ and $j$, the first containing the candidate relation string and the second containing the candidate object string. To avoid examining all possible pairs of fields, we first apply the candidate pair identification algorithm from \citet{Lockard2019OpenCeresWO}, which filters down to a set of potential pairs based on physical and layout distance between text fields.

For each candidate pair, we concatenate the GNN-produced contextual features $h^l$ for both text fields with the original features $h^0$ for both text fields (since some information can be diluted in the GNN), as well as a pairwise feature vector that simply contains the horizontal and vertical distance between the two text fields, and pass them into a binary classifier:
\begin{equation}
r^{OIE}_i  = \text{FNN} \left( [h^0_i; h^0_j;  h_i^l; h_j^l; pairwise_{i,j}], \theta^{OIE} \right)
\end{equation}
\noindent where $\text{FNN}$ is a feed-forward neural network with parameters $\theta^{OIE}$, ``;'' indicates concatenation, and $r^{OIE}_i$ is the predicted probability that the two text fields constitute a \textit{(relation, object)} pair.  
We then optimize for cross-entropy loss across training examples $T$ with $y^{OIE}_i=1$ if the pair is positive:
\vspace{-1em}
\begin{multline}
 \mathcal{L}_{OIE} =  \sum_{i=1}^{T} y^{OIE}_i\log r^{OIE}_i \\ + \left(1-y^{OIE}_i \right)\log \left(1 - r^{OIE}_i \right),
\end{multline}

\subsection{ClosedIE}
\label{sec:closedie_model}
For ClosedIE, we perform a multi-class classification using the contextual representation produced by the GNN ($h^l_i$) along with the original features ($h^0_i$) for text field $i$:
\begin{equation}
r^{CIE}_i  = \text{FNN} \left( [h^0_i; h_i^l], \theta^{CIE} \right)    
\end{equation}

\noindent where $\text{FNN}$ is a feed-forward neural network parameterized by $\theta^{CIE}$, ``;'' indicates concatenation, and $r^{CIE}_i$ is the predicted probability of relation $r$ in set $R$. We optimize for cross entropy loss $\mathcal{L}_{CIE}$:
\begin{equation}
\mathcal{L}_{CIE} = - \sum_{i=1}^{T} \log  p \left( y^{CIE}_i | h^0_i, h^l_i, \theta^{CIE} \right)    
\end{equation}

\noindent where $y^{CIE}_i$ is the true class for example $i$.  For both ClosedIE and OpenIE we use one hidden layer in the feed-forward network.

\section{Experimental Setup}
\label{sec:experimental_setup}

\subsection{Dataset}

For both OpenIE and ClosedIE, our primary dataset is the extended version \cite{Lockard2019OpenCeresWO} of the \textbf{SWDE} dataset \cite{Hao2011FromOT}, which contains gold labels for OpenIE extractions for 21 English-language websites (each with one template) in three subject verticals (Movie, NBA, and University), with between 400 and 2,000 pages per site. We generated ClosedIE labels by converting the OpenIE labels to ClosedIE labels via manual alignment of OpenIE relations between websites, giving a set of 18 relations for the Movie vertical, 14 for NBA, and 13 for University. 
More information on training data creation and a complete listing of ClosedIE relations is available in the Appendix.

We used three SWDE Movie sites (AMCTV, AllMovie, and IMDb) as a development set and did not evaluate on them for the reported results.

\subsection{Experimental Settings}
For each model tested (both our own and the baselines), we classify the training setting into the following categories indicating the level of vertical or site-specific knowledge used, in decreasing level of difficulty. \begin{itemize} [itemsep=1pt,leftmargin=*]
    \item \textbf{Level I--{\newvertical} Zero-shot (OpenIE only):} A model is trained on sites from two of the three verticals (e.g. NBA and University) and applied to sites from the other vertical (Movie). This is the hardest case and is important when we wish to extract knowledge from new verticals where we do not have any prior knowledge or annotations.
    \item \textbf{Level II--Zero-shot with Vertical Knowledge:} A model is trained on all sites but one (spanning Movie, NBA, and University) and then applied to the held-out site. As in cross-validation, experiments are repeated with each site having a turn being held out. It is easier than Level I but is still important for a new website that may not have data overlapping with other websites in the same vertical. For the ClosedIE setting, we train only on in-vertical sites.
    \item \textbf{Level III--Site-specific Knowledge:} This is the traditional setting used by two of our baselines where we have seed knowledge overlapping with the website data to allow training a specific model for the website. Whereas Level I-II are both zero-shot settings, Level III is not, as it allows site-specific training data via weak supervision. (We do not present results using full supervision from manual annotations since it is known from prior work (e.g., \citet{Gulhane2011WebscaleIE}) that full supervision from the target website yields highly accurate semi-structured extractors; we note that {\sysname} also achieves comparable results ($\sim 0.95$ F1) in this setting.
\end{itemize}

We repeated our experiments 10 times and we report the results averaged across the runs. For OpenIE, we follow the ``lenient'' scoring method for SWDE introduced by \citet{Lockard2019OpenCeresWO}, scoring an extraction as correct if the relation string matches any of acceptable surface forms listed by the ground truth for that object.

Models are constructed in PyTorch \cite{paszke2017automatic}, with graph functions implemented in DGL \cite{wang2019dgl} and optimization performed using Adam \cite{Kingma2014AdamAM} and a batch size of 20. For OpenIE, we use a hidden layer size of 25 for the GAT and 100 for the feed-forward layer. For ClosedIE, we use a hidden layer size of 200 for all layers. We use a 2-layer GAT and dropout of 0.25. We obtain visual features by rendering the page using the headless Chrome browser and querying the values using Selenium\footnote{\url{https://www.seleniumhq.org}}.

\vspace{.1cm}
\noindent \textbf{Extraction Threshold:} Since our zero-shot setting means we cannot use a development set of pages from the target site to tune the decision threshold, we instead set the threshold for each experiment to the value that attains the optimal F1 on the experiments where other sites were held-out. 

\vspace{.1cm}
\noindent \textbf{OpenIE Postprocessing Rules:} To ensure consistency among the extracted values, we keep only the highest confidence extraction in the case that the same text field is extracted as both a relation and object, or if multiple relations are extracted for the same object. In addition, some pages in the dataset contain relational tables, from which we sometimes extract the column headers as relations with the column contents as objects. While we believe a post-processing step could potentially recover these relational contents from our extractions, the SWDE data does not contain ground truth for such facts. Instead, we apply the heuristics described by \cite{Cafarella2008UncoveringTR} to identify these tables and remove them from our extractions.

\subsection{Baselines and Models}

We compare against several baselines:

    \vspace{.1cm} \noindent \textbf{Colon Baseline (OpenIE)} This is a heuristic technique that identifies all text fields ending in a colon (``:'') and assumes they are relation strings, then extracts the text field to the right or below, whichever is closer, as the object. We consider it as Level I knowledge since it requires no training. 
    
     \vspace{.1cm} \noindent \textbf{WEIR (OpenIE)}  
     This approach by \citet{Bronzi2013ExtractionAI} discovers relations by aligning multiple pages about the same entity. Because it requires sites to be grouped by vertical and uses a gazetteer list of entity names for the alignment, it has Level III knowledge. 
    
     \vspace{.1cm} \noindent\textbf{OpenCeres (OpenIE)} This applies the model by \citet{Lockard2019OpenCeresWO}, which requires a knowledge base matching some facts presented on the target website, using Level III knowledge.
     
     \vspace{.1cm} \noindent\textbf{\sysname-FFNN (Feed-forward neural network):} This model takes the same features and training data as the full {\sysname} model but removes the GNN component, with versions tested with both Level I (\sysname-FFNN {\newvertical}) and Level II (\sysname-FFNN {\newwebsite}) knowledge.

    \vspace{.1cm} \noindent \textbf{\sysname-GNN}: This applies the full model described in Section \ref{sec:gnn}, with versions tested with both Level I (\sysname-GNN {\newvertical}) and Level II (\sysname-GNN {\newwebsite}) knowledge.

\section{Experimental Results}
\label{sec:results}

\begin{table*}[t]
\renewcommand{\arraystretch}{1}
\centering
\resizebox{\linewidth}{!}{%
\begin{tabular}{lccrrrrrrrrrr} 
\toprule
\multirow{2}{*}{\textbf{System}} & \textbf{Site-specific} & \multirow{2}{*}{\textbf{Level}} & \multicolumn{3}{c}{\textbf{Movie}} &  \multicolumn{3}{c}{\textbf{NBA}} &  \multicolumn{3}{c}{\textbf{University}} & \textbf{Average}  \\
\cmidrule(lr){4-6}
\cmidrule(lr){7-9}
\cmidrule(lr){10-12}
\cmidrule(lr){13-13}
& \textbf{Model} &  & P & R & F1 & P & R & F1 & P & R & F1 & F1\\
\midrule
OpenCeres & Yes & III & 0.71 & 0.84 & \textbf{0.77} & 0.74 & 0.48 & \textbf{0.58} & 0.65 & 0.29 & \textbf{0.40} & \textbf{0.58}\\
WEIR & Yes & III & 0.14 & 0.10 & 0.12 & 0.08 & 0.17 & 0.11 & 0.13 & 0.18 & 0.15  & 0.13\\
\midrule
{\sysname}-FFNN {\newwebsite} & No & II & 0.37 & 0.5 & 0.45 & 0.35 & 0.49 & 0.41 & 0.47 & 0.59 & \textbf{0.52} & 0.46\\
{\sysname}-GNN {\newwebsite} & No & II & 0.49 & 0.51 & \textbf{0.50} & 0.47 & 0.39 & \textbf{0.42} & 0.50 & 0.49 & 0.50 & \textbf{0.47}\\
\midrule
Colon Baseline & No & I & 0.47 & 0.19 & 0.27 & 0.51 & 0.33 & 0.40 & 0.46 & 0.31 & 0.37 & 0.35\\
{\sysname}-FFNN {\newvertical} & No & I & 0.42 & 0.38 & 0.40 & 0.44 & 0.46 & 0.45 & 0.50 & 0.45 & \textbf{0.48} & 0.44\\
{\sysname}-GNN  {\newvertical} & No & I & 0.43 & 0.42 & \textbf{0.42} & 0.48 & 0.49 & \textbf{0.48} & 0.49 & 0.45 & 0.47 & \textbf{0.46}\\
\bottomrule
\end{tabular}}
\caption{With no vertical knowledge, {\sysname}-GNN achieves 65\% higher recall and comparable precision in all verticals compared to the colon baseline. Even in comparison to approaches that use vertical knowledge to learn site-specific OpenIE models, {\sysname} achieves an F1 seven points higher in the University vertical.}
\label{table:swde_results_pr_all_domain}
\end{table*}

\subsection{OpenIE}

\noindent{\bf Level-I Knowledge:} Table \ref{table:swde_results_pr_all_domain} shows that {\sysname} is able to extract facts in entirely new subject verticals 31\% more accurately than the colon baseline. Across all SWDE sites (micro-averaging across all extractions), \sysname-GNN achieves an F1 of 0.45, in comparison with 0.43 for \sysname-FFNN, showing that the additional information provided by the page encoder allows for a better representation of the relationships between text fields.

By successfully learning general patterns of relational presentation on webpages, \sysname-GNN is able to train solely on a set of 16 websites about Movies and NBA players, and then extract from University websites more accurately than the WEIR and OpenCeres systems, which take advantage of Level III knowledge to learn models specific to those University sites. While OpenCeres's rich vertical knowledge allows it to attain better results in Movie and NBA, \sysname-GNN still posts much stronger results than the other baselines in these two verticals. 
 
\begin{figure}[t]
  \centering
    \includegraphics[width=1.0\linewidth]{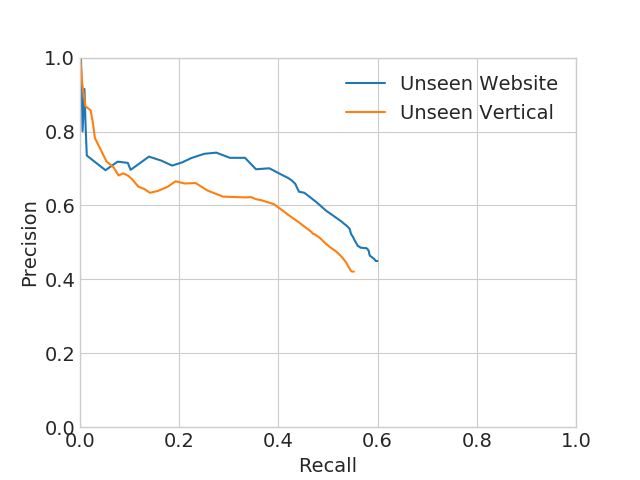}
  \caption{For OpenIE, using the full SWDE set (except the test site), including in-vertical training data (i.e. Level II knowledge), allows for 5-10 point gains in precision at equivalent recall compared to using only out-of-vertical training data (Level I).}
  \label{fig:inVoutDomain}
  \vspace{-0.5em}
\end{figure}

\smallskip
\noindent{\bf Level-II Knowledge:}
Figure \ref{fig:inVoutDomain} shows that adding the in-vertical sites to the training set (but still withholding the test site) allows the model to achieve performance better than the Level I training set that uses only out-of-vertical data.

\begin{figure}[t]
  \centering
    \includegraphics[width=1.0\linewidth]{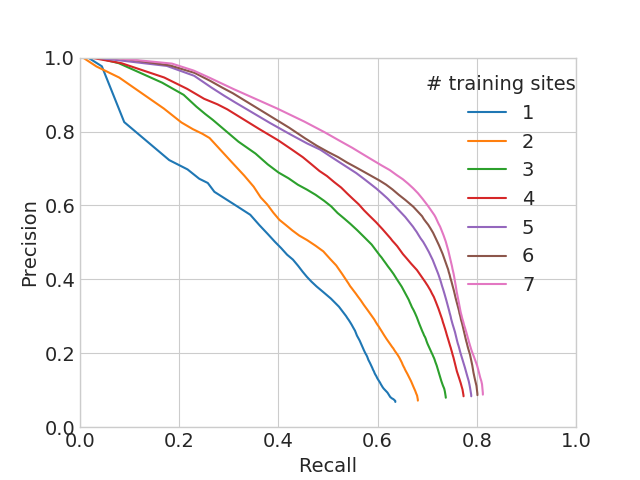}
  \caption{Performance on the ClosedIE Movie vertical increases significantly as more sites are added to the training data.}
  \label{fig:varying_music}
  \vspace{-0.5em}
\end{figure}

\subsection{ClosedIE}

\begin{table}[t]
\renewcommand{\arraystretch}{1}
\centering
\resizebox{\linewidth}{!}{%
\begin{tabular}{lcrrr} 
\toprule
System & Knowledge Level & P & R & F1 \\
\midrule
{\sysname}-FFNN & II & 0.45 & 0.49 & 0.46 \\
{\sysname}-GNN & II & 0.62 & 0.55 & \textbf{0.58}  \\
\bottomrule
\end{tabular}}
\caption{For ClosedIE, using the pre-trained GNN adds 12 F1 points in comparison to the baseline lacking contextual information.} 
\label{table:swde_results_closed_pr}
\vspace{-1em}
\end{table}

\begin{table}[t]
\renewcommand{\arraystretch}{1}
\centering
\resizebox{0.9\linewidth}{!}{%
\begin{tabular}{lrr} 
\toprule
 & OpenIE F1 & ClosedIE F1 \\
\midrule
Full Model & 0.71 & 0.73\\
\midrule
No GNN & 0.68 (0.03 $\downarrow$ ) & 0.63 (0.10 $\downarrow$)\\
No pre-training & 0.66 (0.05 $\downarrow$) & 0.73\\
No DOM edges & 0.65 (0.06 $\downarrow$) & 0.58 (0.15 $\downarrow$)\\
No spatial edges & 0.65 (0.06 $\downarrow$) & 0.62 (0.11 $\downarrow)$\\
No visual features & 0.55 (0.16 $\downarrow$) & 0.73\\
No BERT features & -- & 0.10 (0.63 $\downarrow$)\\
Add BERT features & 0.68 (0.03 $\downarrow$) & --\\
\bottomrule
\end{tabular}}
\caption{Ablations on the Movie development set.} 
\label{table:movie_ablations}
\vspace{-1em}
\end{table}

Table \ref{table:swde_results_closed_pr} shows the results for ClosedIE extraction. \sysname-GNN attains an overall F1 of 0.58 averaged across the three verticals. This significantly outperforms the feed-forward model that did not use the GNN, which attained an F1 of 0.46. While our performance on this dataset is far below the state-of-the-art for semi-structured ClosedIE (above 0.9 for all verticals), prior systems all learn site-specific models based on manual labeling or prior knowledge aligned to the website, while we have only Level II Knowledge available.

Figure \ref{fig:varying_music} shows how adding additional training data improves performance in the Movie vertical. It appears that adding additional training sites would further improve the performance.

\subsection{Ablation Study}

Table \ref{table:movie_ablations} shows the contributions of different elements of the model in the OpenIE and ClosedIE settings as calculated on the development set of three sites in the Movie vertical. These ablations show that the GNN helps in both settings, with a larger effect in ClosedIE, which is likely due to sharing the rich information about the text of nearby text fields.

Pre-training is important in OpenIE but does not have a significant effect for ClosedIE. This is not surprising given that the pre-training task is closely related to the OpenIE task. Both DOM and spatial adjacency edges contribute to the success of the page layout graph for the GNN. In the ClosedIE setting, the text and layout relationships alone will generally contain sufficient information to make an extraction, while in OpenIE the visual elements (such as whether text is bold or underlined) are a strong source of consistency across websites.

\subsection{Error Analysis}
\noindent\textbf{OpenIE:} 
To understand what cases our \sysname-GNN model is missing, we sampled 100 error cases in each vertical from the {\newvertical} experiment and manually examined them. Some examples of both erroneous and correct extractions are shown in Table \ref{table:errors} in the Appendix.
False positives were largely due to the presence of two different types of n-ary relationships on the page.

The first class of errors involving n-ary relationships, making up 43\% of all false positives, were where several facts have a multi-way relationship with the page topic, but individually the fields are not meaningful. For example, the NBA site USAToday includes a ``Latest notes'' section with links to several articles relevant to the page topic entity, mentioning the date, headline, and summary. We extract all of these objects with the ``Latest notes'' relation, but to obtain meaningful knowledge it would be necessary to additionally associate the correct date, headline, and summary with each other. While we can envision methods for doing this via post-processing, the SWDE benchmark considers these to be errors.

In the second class, {\sysname} correctly extracted \textit{(relation, object)} pairs, but from page sections that contain facts about entities other than the page topic. For example, on the MatchCollege site, a section of ``Similar Local Colleges'' contains some of the same relations presented for the page topic, in similar formatting. These types of errors made up another 6\% of false positives.

Of the remaining errors, 33\% were due to the extraction of pairs where the extracted relation did not represent a relationship, while another 14\% were due to the extraction of pairs with a correct relation string and incorrect object. Most false negatives occurred in long vertical lists, where some values were extracted, but not all.

\smallskip
\noindent \textbf{ClosedIE}: False negatives were most likely to occur on long lists of values (such as cast lists), where values toward the bottom of the list were sometimes missed. Recall also suffered on relations where the relation name varied significantly from site to site, or where ambiguity existed. For example, the string ``Produced by'' is used by some sites to indicate the producer of the film, while on other sites it indicates the production company.

\section{Conclusion}
\label{sec:conclusion}
We have introduced a zero-shot method for learning a model for relation extraction from semi-structured documents that generalizes beyond a single document template. Moreover, this approach enables OpenIE extraction from entirely new subject verticals where no prior knowledge is available. By representing a webpage as a graph defined by layout relationship between text fields, with text fields associated with both visual and textual features, we attain a 31\% improvement over the baseline for new-vertical OpenIE extraction. Future extensions of this work involve a more general pre-training objective allowing for the learned representations to be useful in many tasks as well as distantly or semi-supervised approaches to benefit from more data.

\section*{Acknowledgments}
We would like to acknowledge grants from ONR N00014- 18-1-2826, DARPA N66001-19-2-403, NSF (IIS1616112, IIS1252835),  Allen Distinguished Investigator Award, and Sloan Fellowship.

\bibliography{anthology,acl2020}
\bibliographystyle{acl_natbib}

\newpage
\appendix

\section{Appendix}

\begin{table*}[!t]
\renewcommand{\arraystretch}{1.2}
\normalsize
\centering
\resizebox{1.0\linewidth}{!}{%
\begin{tabular}{llllp{3cm}cp{4cm}} 
\toprule
\multirow{2}{*}{\textbf{Vertical}} & \multirow{2}{*}{\textbf{Site}} &  \multicolumn{3}{c}{\textbf{Extraction}} &  \multirow{2}{*}{\textbf{Correct}} &  \multirow{2}{*}{\textbf{Notes}}  \\
\cmidrule(lr){3-5}
& & Page Topic & Relation & Object & &  \\
\midrule
Movie & Hollywood & Spanish Fly & Costume Designer & Jose Maria de Cossio & Yes &  \\
\midrule
Movie & Metacritic & Saving Face & Reviewed by & Maitland McDonagh & Yes &  \\
\midrule
NBAPlayer & ESPN & Jameer Nelson & Birth Place & Chester, PA & Yes &  \\
\midrule
NBAPlayer & MSNCA & Matt Bonner & College & Florida & Yes &  \\
\midrule
University & CollegeProwler & Spring Arbor University & Admission Difficulty & Average & Yes &  \\
\midrule
University & MatchCollege & Menlo College & College Credits Accepted & AP Credit & Yes &  \\
\bottomrule
Movie & RottenTomatoes & Slow Burn & Tomatometer Percentage & 97\% & No & Subject of relation is not page topic but is an unrelated recently released film \\
\midrule
Movie & RottenTomatoes & Ginger Snaps 2 & WHAT'S HOT ON RT & Trailer: Santa has a bloody Xmas & No & Extracted relation string is not a relation \\
\midrule
Movie & Metacritic & The Constant Gardener & User Panel Options & The Constant Gardener & No & Extracted relation string is not a relation \\
\midrule
University & CollegeProwler & Minnesota School of Business & CP Top 10 Lists & Best Performance Venues & No & Link to article not related to page topic, but is a ``Similar School'' \\
\midrule
University & MatchCollege & Maric College & Highest Degree & Associate's & No & Subject of relation is not page topic \\
\midrule
NBAPlayer & FoxSports & Tony Parker & Latest News & Mon. Dec 6, 2010 & No & n-ary object \\
\midrule
NBAPlayer & MSNCA & Gilbert Arenas & Birthplace & 215 & No & Erroneous extraction of weight for birthplace (both text fields are nearby) \\
\bottomrule
\end{tabular}}
\caption{Selected OpenIE Extractions from \sysname-GNN with Level I training (no knowledge of the subject vertical).} 
\label{table:errors}
\end{table*}

\subsection{ClosedIE Label Mappings}
SWDE provides OpenIE labels for all binary relations between the objects mentioned on the page and the page topic entity. These labels include the relation string used to indicate the relationship, sometimes including multiple acceptable surface forms if there is more than one applicable string for the relation (usually due to more or less specific versions of the relation). The original SWDE data only includes ClosedIE labels for a small subset of relation types. To create ClosedIE ground truth for all relations on the sites, we examined all OpenIE relations across the SWDE sites and grouped them into a set of relations that each represented the same fundamental idea. In some cases, we chose to map relations into a somewhat more general category, such as mapping ``Associate Producer'' and ``Executive Producer'' into the same ``Producer'' concept. After obtaining this set, we eliminated all relations that appeared on fewer than 3 websites in the dataset. The set of relations used for the ClosedIE experiments is given in Table \ref{table:label_mappings}. The full mapping of OpenIE to ClosedIE relations can be found at \url{https://github.com/cdlockard/expanded_swde}.

\begin{table*}[t]
\renewcommand{\arraystretch}{1.2}
\centering
\resizebox{0.5\linewidth}{!}{
\begin{tabular}{lr} 
\toprule
Vertical & Relation\\
\midrule
movie & movie.aka \\
movie & movie.box\_office \\
movie & movie.budget \\
movie & movie.country \\
movie & movie.directed\_by \\
movie & movie.distributor \\
movie & movie.genre \\
movie & movie.language \\
movie & movie.produced\_by \\
movie & movie.production\_company \\
movie & movie.rating \\
movie & movie.release\_date \\
movie & movie.runtime \\
movie & movie.starring \\
movie & movie.synopsis \\
movie & movie.written\_by \\
movie & movie.year \\
nbaplayer & nbaplayer.age \\
nbaplayer & nbaplayer.assists \\
nbaplayer & nbaplayer.birthdate \\
nbaplayer & nbaplayer.birthplace \\
nbaplayer & nbaplayer.college \\
nbaplayer & nbaplayer.draft \\
nbaplayer & nbaplayer.experience \\
nbaplayer & nbaplayer.field\_goal\_percentage \\
nbaplayer & nbaplayer.height \\
nbaplayer & nbaplayer.points \\
nbaplayer & nbaplayer.position \\
nbaplayer & nbaplayer.rebounds \\
nbaplayer & nbaplayer.weight \\
university & university.application\_fee \\
university & university.calendar\_system \\
university & university.control \\
university & university.enrollment \\
university & university.in\_state\_tuition \\
university & university.out\_state\_tuition \\
university & university.phone \\
university & university.religious\_affiliation \\
university & university.setting \\
university & university.tuition \\
university & university.undergraduate\_enrollment \\
university & university.website \\
\bottomrule
\end{tabular}}
\caption{A listing of ClosedIE relation types mapped from OpenIE labels in SWDE}
\label{table:label_mappings}
\end{table*}

\subsection{Training Data Creation}
\label{app:training_data_creation}
The Extended SWDE dataset provides ground truth extractions of OpenIE predicate and object strings for the webpages it contains. However, it does not specify which text fields on the page were the source of the extractions. To create training data, we need to label a specific text field. It is usually the case that each ground truth string matches only one text field, so there is no ambiguity, but in cases where multiple text fields have the same value, we must disambiguate which one to use. We did this by identifying all matching text fields for the ground truth predicate and object and chose the pair in which the predicate and object strings have the closest Euclidean distance on the rendered page.

While this is generally a safe assumption, there are still occasional errors in the training data. In particular, we observed that the NBA vertical had considerably more ambiguous cases since most relations are numerical and the pages often contained large tables of numbers. We hypothesize that this may explain why performance on the NBA vertical is lower when using {\newwebsite} training data compared to the {\newvertical} setting (Table \ref{table:swde_results_pr_all_domain}).

During testing, we applied the same standard used by prior work on the dataset and accepted an answer as correct if it matched the ground truth string, regardless of which text field produced the extraction.

\end{document}